# Fixing the Twin, Not the Detector: Sim-to-Real Roadside LiDAR Detection from Digital-Twin Data

KuanLin Chen[1,2,*] Feng-jui Kuo[2]
[1]*Department of Industrial and Systems Engineering, Chung Yuan Christian University, Taoyuan, Taiwan*
[2]*Tacitt, Karma Laboratories, Inc., Baltimore, MD, USA*
s11224246@cycu.edu.tw, raykuo@karmalabs.ai
[*]*Corresponding author*

***Abstract*—Roadside LiDAR detectors are expensive to train on real labels, and digital twins promise to replace these labels, but a detector trained on a twin meets a different sensor, different objects and different annotation conventions on the real site. We describe the pipeline that placed first in the LUMPI track of the UrbanTwin Sim2Real LiDAR Challenge (IEEE BigData 2026 Cup, 0.5541 combined score) and fifth on its V2X-Real track (0.8102). Every detector is a CenterPoint model trained only on synthetic data, on a single 8 GB laptop GPU. Most of the gain came from correcting differences between the twin and the real site rather than from the detector. We align the twin's coordinates, point density and box convention with the test sensor, generate the pedestrians and riders that the twins lack, plus street clutter as unlabeled negatives, and rebuild twin objects at real size and real point density. A non-learned resampling to the per-cell point density of unlabeled real frames lifts LUMPI realism from 0.589 to 0.947, against 0.949 between real frames. Label-free post-processing then exploits what a fixed roadside sensor guarantees: static structures recur in every frame, and real objects have characteristic point statistics. Reporting the validation and test gain of each component side by side, we find that corrections of systematic sim-to-real differences transfer to the test set, typically keeping a quarter to three quarters of their validation gain, while choices that fit the few objects of a validation set do not.**

## I. Introduction

Roadside LiDAR sensors watch intersections from a fixed pose and feed traffic monitoring and cooperative driving. Training their 3D detectors needs labeled frames from each site, which are slow and costly to collect. Digital twins of the sites, rendered in a driving simulator, could supply unlimited labeled data, but a detector trained on a twin meets a different sensor pattern, different objects and different annotation conventions on the real site.

The UrbanTwin Sim2Real LiDAR Challenge [1] tests this setting directly. Detectors may be trained only on synthetic data and are scored on real frames of two intersections: LUMPI in Hannover [2] and V2X-Real in Los Angeles [3]. Each submission holds 50 synthetic frames, scored for realism against hidden real frames, and detections on 50 released real test frames, scored by 3D mean average precision (mAP). The organizers provide UrbanTwin digital twins of both sites [4]–[6].

We ask three questions:

**RQ1.** Which differences between a digital twin and the real site limit a detector trained only on the twin, and how much of this gap can be closed by correcting the synthetic data, with real data used only for statistics and validation? (Sections IV and VII)

**RQ2.** What can label-free post-processing gain from the fixed pose of a roadside sensor, and where does it stop? (Sections VI and VII)

**RQ3.** When real frames may be used only for validation, which validation gains carry over to the test set? (Sections VII and IX)

We found that most of the achievable gain lies in the twin itself. Four differences between twin and test data decided the score: the twins contain no walking pedestrians; twin frames carry about 220,000 points while test frames are cropped to 50,000; twin objects are smaller and sparser than real objects, and their boxes are tighter than real annotations; and the LUMPI twin is offset from the real sensor frame. Correcting these at the source, in the training and realism data, moved the score more than any change to the detector or its post-processing.

Our contributions answer these questions:

1. Source-level correction of the twin, guided only by statistics of real labels and unlabeled real frames: twin frames are aligned with the test sensor in coordinates, density and box convention; objects are rebuilt at real median size and real point density, extending size normalization [7] to point density; and the pedestrians, riders and street clutter that the twins lack are synthesized procedurally, the clutter as unlabeled negatives.
2. A realism generator without any learned model: exact per-cell density matching to unlabeled real frames lifts the LUMPI realism score from 0.589 to 0.947, against 0.949 between real frames, and is the largest gain of our pipeline (+0.14 combined).

3. Label-free post-processing that exploits a fixed sensor: priors for structures that recur in every frame, point-feature rules for real objects, and point-fitted box geometry.
4. A validation protocol on non-forbidden real frames, with candidates and thresholds fixed before evaluation in the final selection rounds, and an ablation that compares validation and test gains. Corrections of systematic sim-to-real differences typically kept a quarter to three quarters of their validation gain on test, and the final LUMPI round, validated on in-window frames recorded in the same sessions as the test frames, kept about 85%. Choices fitted to the few objects of a validation set lost on test.

The pipeline (Fig. 1) placed first in the LUMPI track (0.5541 combined score) and fifth on V2X-Real (0.8102), with every model trained on a single 8 GB laptop GPU. Section IX distills six take-aways for training detectors on digital twins.

## II. Related work

**LiDAR simulation.** Driving simulators such as CARLA [8] render labeled LiDAR scans of virtual scenes, and the UrbanTwin datasets [4], [5] use it to reproduce the LUMPI and V2X-Real intersections. LiDARsim [9] narrows the gap by building assets from real scans and learning the sensor's dropout, and domain randomization [10] trains on wide synthetic variation so that the real data looks like one more variation. The challenge allows neither real scans in training nor a learned real sensor model, so we correct the synthetic data with a few statistics of real data instead.

**Cross-domain 3D detection.** Wang et al. [7] showed that the gap between driving datasets is dominated by object size, and that rescaling source objects to the target's mean size recovers much of it. Self-training on unlabeled target frames, as in ST3D [11], is the other main family, but it puts real frames into training and is excluded here. Our real-size rebuilding of twin objects applies the insight of [7] to a sim-to-real setting and extends it to point density.

**Roadside detection.** Fixed infrastructure sensors appear in LUMPI [2], V2X-Real [3] and DAIR-V2X [12]. We build on the voxel detectors of OpenPCDet [13]: CenterPoint [14] as the base model, ground-truth sampling from SECOND [15], one Voxel R-CNN [16] member, and weighted box fusion [17] to combine members. A fixed pose makes the background repeat in every frame, which our site priors exploit.

**Point-cloud realism.** The challenge scores the synthetic frames with the Chamfer distance (CD) and earth mover's distance (EMD) [18], the Fréchet point-cloud distance (FPD) [19] and maximum mean discrepancy (MMD) [20] between pooled synthetic and real points. These compare coarse 3D occupancy, not individual objects, which is why per-cell density matching is effective.

**Earlier solutions.** At an earlier edition of this challenge, held with the DriveX workshop at ECCV 2026, Luo et al. [21], [22] aligned the synthetic scan density with the test data, combined transformer- and pillar-based detectors through class-aware fusion, and built the realism frames by matching the radial point density. We match the density per 3D cell instead, which reaches the real-versus-real ceiling, and also correct object size, object density and the missing classes at the source.

## III. Challenge, data and metrics

Each track has its own synthetic training data, real test frames and scored classes (Table I). The 50 test frames of a track are real, unlabeled LiDAR frames; predictions are scored by 3D mAP, the mean over classes of the average precision (AP), at an intersection-over-union (IoU) of 0.5 for every class, with the 40-point interpolated AP [23] used by the KITTI benchmark [24] and without difficulty filtering. The 50 submitted synthetic frames are compared with 50 hidden real frames after pooling and subsampling to 10,000 points, using the four realism metrics of Section II, each normalized to [0, 1]. The combined score is $0.6 \times \min(1, \text{mAP} / 0.7) + 0.4 \times$ the mean normalized realism. The challenge had a development and a final phase with separate leaderboards, both scored on the same 50 test frames.

TABLE I
Tracks, scored classes and evaluation ranges

| Track | Real site | Scored classes | Evaluation range (m) |
|---|---|---|---|
| LUMPI | Hannover intersection, 5 fused roadside LiDARs | Person, Car, Bicycle, Motorcycle, Bus, Truck, Van | [−40, 40] × [−40, 40] × [−5, 5] |
| V2X-Real | UCLA intersection, infrastructure LiDAR | vehicle, pedestrian, truck (Van and Bus count as truck, riders as pedestrian) | [−40, 40] × [−40, 40] × [−8, 2] |

Three rules shaped the design. Detectors may be trained or fine-tuned only on synthetic data. The 100 frames listed as forbidden, which include the test frames, may not be used for training, validation, tuning or inspection of ground truth. The remaining public real data may be used for validation, model selection and box-size statistics, unlabeled real point clouds may be used to calibrate the synthetic data, and every use of real data must be disclosed (Section VIII).

The UrbanTwin data provide 9,000 synthetic frames per track (UT-LUMPI [25] and UT-V2X-Real-IC [26]) with seven-parameter boxes (center, size, heading). The four differences named in Section I look as follows. The twins contain no walking pedestrians, and UT-LUMPI has no Person labels at all. Twin frames carry about 220,000 points, while test frames are cropped to 50,000. Twin objects are smaller and sparser than real ones, and twin boxes are tighter than the real annotation convention (a real Bicycle box includes its rider). The LUMPI twin is offset from the real sensor frame by (−1.25, 0.25, 0.04) m.

## IV. Synthetic training and realism data

### A. Training frames

Each twin frame is translated into the real sensor frame, cropped to the evaluation range and randomly subsampled to 50,000 points, so the detector only ever sees the test density. Labels are mapped to the track's classes, and boxes with fewer than 5 points are dropped. Because twin boxes are smaller than the real annotation convention, the LUMPI Bicycle, Motorcycle, Van and Bus training boxes are rescaled toward the size statistics of non-forbidden real labels; predicted Car boxes are set to the real median size at inference (Section VI).

### B. Procedural pedestrians, riders and clutter

Neither twin contains walking pedestrians, so we synthesize them inside the data loader. A person is sampled on capsule and ellipsoid primitives (height 1.6 to 1.9 m, with a walking stride), kept only on the side facing the nearest sensors, and thinned to a point count drawn from a log-normal model of points versus distance, calibrated to real pedestrian point counts (LUMPI: median 70 points at 10 m). People are placed on open ground and receive a Person box; a crowd variant places groups of 2 to 5 people with mutual occlusion. On the labeled V2X-Real validation frames, two thirds of the confident pedestrian boxes sat on unlabeled low clusters (bollards, bins, sign posts, shrubs, parked bikes) that the twins lack, so we add eight kinds of procedural street clutter as unlabeled background with the same point model. For V2X-Real, where riders count as pedestrians, one member was further fine-tuned with procedural riders (a bicycle, scooter or motorcycle carrying a seated person), boxed at the median size of real rider labels.

### C. Scene augmentations

Three augmentations target the remaining differences: random point dropout (keep ratio U[0.4, 1] or U[0.3, 1]) for distant vehicles with fewer than 20 points; a whole-scene z shift after ground-truth sampling (U(−0.2, 0.8) m on V2X-Real, U(−0.6, 0.2) m on LUMPI), because the real ground does not lie at the synthetic height; and box occlusion, which drops the points in an azimuth window covering 20 to 60% of a box or thins them with range.

### D. Real-size, real-density objects

Twin objects are both too small and too sparse (Table II). At 50,000 points the twin's cars extend 3.4 × 1.4 m and reach 1.2 m above the ground, against 4.0 × 1.8 m and 1.5 m for real Car labels, and real objects of every class carry about twice the points at the same range. On V2X-Real the gap is largest for trucks. For the final members we therefore rebuilt the training frames from the raw twin frames. Object points and boxes are scaled per axis about the box bottom center to the real median size, unless the enlarged footprint would touch another box (3% of the LUMPI cars); before the frame is cut to 50,000 points, the points of every labeled object are kept with 2.3 times the keep probability of the other points. Fine-tuning on these frames raised raw single-member Car AP on the in-window LUMPI frames (M4w, Section VII-A) from 0.557, 0.562 and 0.570 to 0.623, 0.629 and 0.636, and Bicycle AP from 0.044 and 0.035 to 0.074 and 0.082. On V2X-Real, the first of the two truck members rebuilt this way raised second-half truck AP by 0.064, the largest single gain of the final V2X-Real rounds (Section VII-E); the same recipe did not help vehicles, whose twin and real sizes differ less. Thinning the test frames toward the twin's density instead only lost accuracy.

TABLE II
Twin versus real object statistics

| Quantity | Twin | Real |
|---|---|---|
| LUMPI Car point extent at 50k points (m) | 3.4 × 1.4, top 1.2 | 4.0 × 1.8, top 1.5 |
| LUMPI Car points per 10 m range bin | 213, 231, 107, 38 | 406, 456, 213, 70 |
| V2X-Real truck size (m, median) | 5.2 × 2.1 × 2.1 | 7.8 × 2.8 × 2.6 |

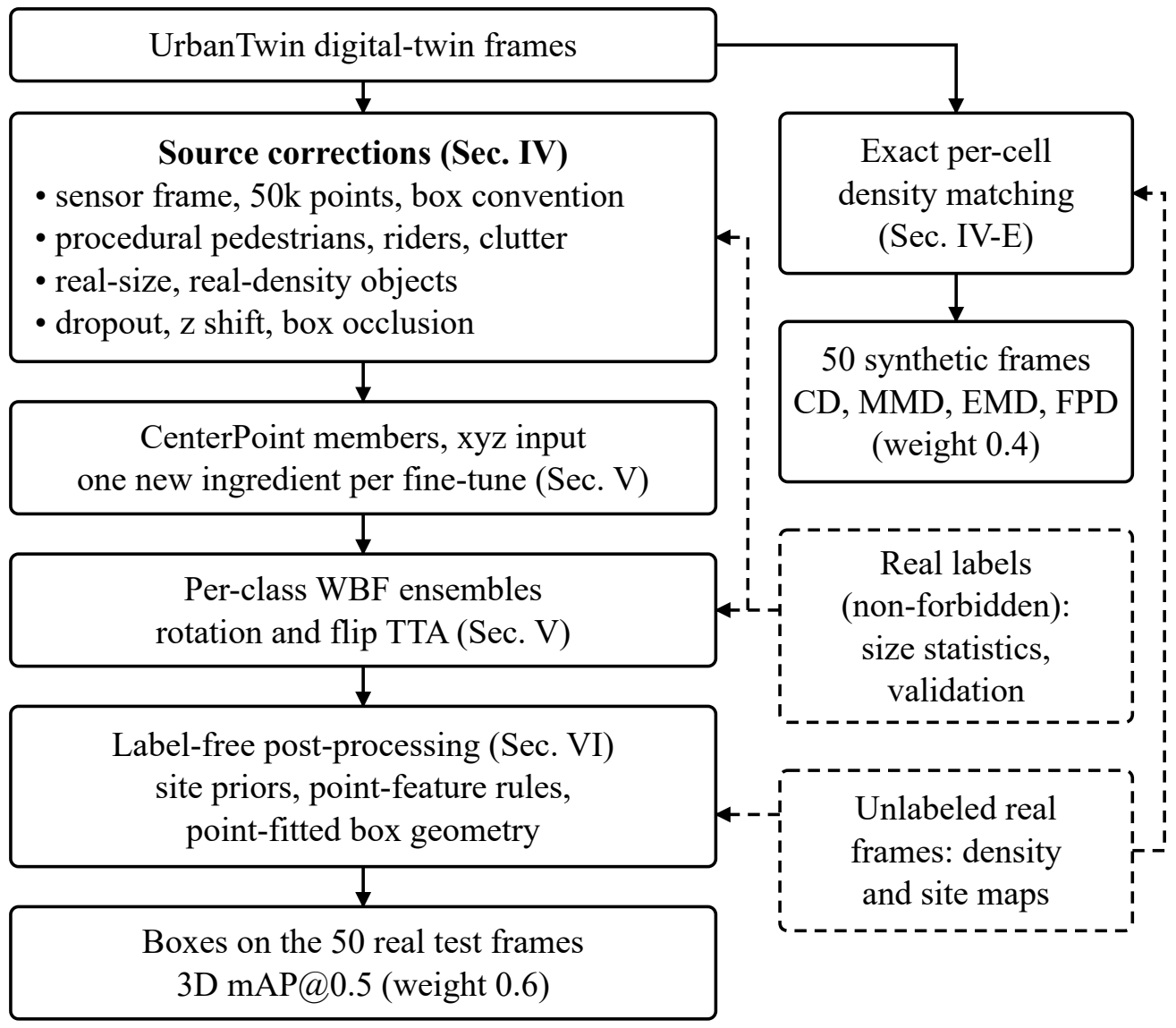

Fig. 1. Pipeline. Twin frames are corrected at the source (coordinates, density, box convention, missing classes, real-size and real-density objects) before any training; per-class ensembles with test-time augmentation produce boxes, and label-free post-processing uses only the test frame's own points and maps from unlabeled real frames. Dashed: real data from non-forbidden frames; labels give size statistics and validation, unlabeled frames give the density and site maps.

### E. Realism frames

The realism metrics pool the 50 synthetic frames, so coarse 3D occupancy matters, not individual objects. Raw twin frames score 0.589 on LUMPI because the twin places its points differently from the real sensors. Each submitted frame is built by per-cell density matching: the evaluation range is divided into 1 m × 1 m columns with 25 height layers, and the target count of a cell is the point budget times the real density of that cell, measured on unlabeled non-forbidden calibration frames. Twin points fill a cell where enough exist, and the re-mainder is drawn uniformly inside it; about 29% of the submitted points are twin points. On held-out check frames this gives realism 0.947 on LUMPI, against a real-versus-real ceiling of 0.949, and 0.951 on V2X-Real; on the leaderboard EMD fell from 2.9 to 1.10 and FPD from 75 to 0.53. Calibrating on the in-window frames of the final LUMPI round (Section VII-D) lowered EMD further to 1.04 and FPD to 0.29.

## V. Detector and ensembles

**Model.** Every member is a CenterPoint detector [14] from OpenPCDet [13]: mean voxel features, a sparse residual 3D backbone (voxels 0.1 × 0.1 × 0.25 m, stride 8; stride 4 for the V2X-Real pedestrian members), a bird's-eye-view (BEV) backbone and one center head per class group; one V2X-Real vehicle member adds a Voxel R-CNN second stage [16]. Input is xyz only, because the test frames carry no intensity. Base models train for 20 to 30 epochs with Adam [27] and a one-cycle schedule [28] (peak learning rate 0.002, batch 4) on one laptop GPU (RTX 5060, 8 GB). Most later members are 8-epoch fine-tunes of their parent that add exactly one synthetic ingredient of Section IV, so each member's contribution can be attributed. A member and its fine-tunes form a lineage, and a package is one complete submission: members, fusion, post-processing and synthetic frames.

**Per-class ensembles.** Members are fused per class by weighted box fusion (WBF) [17]: boxes are clustered greedily by BEV IoU (0.4; 0.6 for V2X-Real vehicles and, in the final package, V2X-Real trucks), and each cluster becomes one box with the score-weighted center and size, the heading of its top-scoring box, and a score equal to the summed member scores divided by the number of runs. A box found by one member only is thus down-weighted, and a weak member dilutes every cluster it misses: removing the oldest V2X-Real vehicle members added 0.0035 vehicle AP on test. Each class comes from its own member set, so a change to one class's members reaches another class only through the cross-class copies and rider boxes of Section VI.

**Test-time augmentation (TTA).** On V2X-Real every member runs with flips (none, x, y, xy), which raised validation mAP from 0.434 to 0.461; on LUMPI flips gained on validation and lost on test. Rotating the scene about the sensor axis by ±5° (V2X-Real also ±10°) and rotating the boxes back transferred on both tracks, with combined test gains of +0.0042 for V2X-Real pedestrians (measured together with a change of the vehicle fusion threshold), +0.0039 for V2X-Real trucks, +0.0022 for LUMPI Person and +0.0039 for LUMPI Car and Truck. A rotation changes how points fall into voxels, which adds diversity that flips do not.

**Member sets.** In the final LUMPI package, Person comes from six z-shift members of the crowd and clutter lineage, each run at 0° and ±5°. Car comes from two real-size members (Section IV-D) at 0° and ±5°, plus the Car boxes of a third real-size member of the Person lineage. Bicycle comes from the same two real-size members; Truck, Bus and Motorcycle come from two base models, Truck and Bus with rotations. Van comes from a base model trained on the twin as released, the best Van source on validation. On V2X-Real, vehicle and truck come from the dropout members and their z-shift and occlusion fine-tunes, and pedestrian from the stride-4 members with crowds and clutter. The final package re-selected each set on in-window frames and added the two real-size truck members, the rider boxes of the rider-trained member and the pedestrian outputs of a real-size vehicle member.

## VI. Label-free post-processing

Post-processing uses only the test frame's points, the detector's boxes and site maps computed from unlabeled non-forbidden frames, and handles each test frame independently; no step uses hand-specified locations. Steps are applied per class and merged, so each step changes only the class it targets. Where a rule multiplies a box by a factor, it scales the box's score. The few hyperparameters were selected on validation frames (Section VII-A).

**Geometric calibration.** Predicted boxes sat systematically high on LUMPI, so each class gets a constant z offset (Car −0.13, Person −0.05, others −0.10 m); Car boxes take the real median size and Person boxes are enlarged by 1.1 × 1.1 in plan. On V2X-Real, truck boxes are scaled by 1.15 in length and 1.08 in width toward the real truck size of Table II, and by 0.95 in height, as chosen on validation; in the final package a heading consensus rotates each fused truck box 25% of the way toward the score-weighted mean heading of its member boxes, with headings taken modulo 180°. Pedestrian and two-wheeler centers are moved halfway toward the centroid of nearby points, and their box bottoms are placed on the local ground (30th percentile height of the points in a ring around the box). Annotators fit boxes to visible points, so box height and center are blended halfway toward the 2nd and 98th percentile heights of the in-box points (LUMPI Car and Bus, V2X-Real truck; center only for V2X-Real pedestrians). Two corrections address the steep viewpoint of the V2X-Real sensor: a near-range z ramp lowers pedestrian boxes at range $r < 10$ m by $0.5 \times (1 - r/10)$ m (+0.010 test AP), and the pedestrian members run on frames shifted 0.5 m down in z, with their boxes shifted back (+0.008 test AP).

**Site priors.** Both sensors are fixed, so false positives on poles, bushes and sign posts recur at the same place in almost every frame, while road users move. On LUMPI validation, 60% of the Person false positives sat on cells that held a Person prediction in at least half of all frames, against 19% of the true detections. From unlabeled real frames (LUMPI: 440 frames at least 300 frames from any forbidden id; V2X-Real: the validation frames, leave-one-scene-out) we record, per 0.5 m cell and class, the share of frames with a prediction of score at least 0.3 in the 3 × 3 neighborhood; boxes on cells with a share of at least 0.5 are multiplied by 0.3 (V2X-Real 0.5). A second map records, per 0.3 m voxel, the share of frames in which the voxel holds a point; a Person, Bicycle or Motorcycle box whose above-ground points lie on average in voxels occupied in at least 40% of the frames is multiplied by 0.3.

**Point-feature rules.** More than half of our LUMPI Bicycle false positives sat on pedestrians, so Bicycle and Motorcycle boxes whose above-ground points extend less than 0.7 m along the heading are multiplied by 0.5. For small-object boxes we take the body points, those within 0.45 m of the center and more than 0.3 m above the box bottom, and record their count and top: V2X-Real pedestrians with fewer than 6 such points or a top below 1.2 m are multiplied by 0.3 (true pedestrians have about 16 points and reach 1.5 m; background clusters have 5 points and reach 0.86 m). LUMPI uses the same rule with class-specific thresholds. The final V2X-Real package rescales the pedestrian scores with a smooth version of these features (points inside the box, occupied height bands, agreement with the local ground and the point count expected at the box's range), with parameters chosen by random search on validation frames.

**Cross-class copies.** About 10% of the Car labels in the LUMPI validation sets were matched only by a Van, Bus or Truck box. AP is computed per class and low-scoring boxes cost almost nothing, so every Van box is also added to the Car list at 0.7 of its score and every Bus and Truck box at 0.3 (+0.034 test Car AP), with smaller copies into Truck and Van.

**Two-wheeler and wide-object rules (final LUMPI round).** Four class-level rules respond to the M4w label statistics of Section VII-D. (i) Two-wheeler refit: every Bicycle and Motorcycle box and every long Person box takes the footprint 1.95 × 0.86 m, the median two-wheeler label of non-forbidden real frames; where the point cluster under the box is elongated (1.0 to 2.6 m along its main axis and at least 1.4 times as long as wide), the box takes the heading of that axis and the center of the cluster. (ii) Rider as Person: two-wheeler boxes with score at least 0.1 are grouped into objects (centers within 0.6 m). Each object also outputs its best box as a Person box (score mapping chosen on M4w, Section VII-D); this box replaces lower-scoring long Person boxes nearby. Short Person boxes within 0.6 m of an object that scores at least 0.3 are multiplied by 0.3. (iii) Wide objects: for each Truck or Bus box we take the connected blob of above-ground points under it; if the minimum-area rectangle of its points at least 2 m above ground is at least 4.0 m wide, that rectangle is added as a Truck box at half the score. (iv) Every Truck box shorter than 7.5 m is also added as a Van box with the same score, unless a Van box within 1.0 m already scores as much.

**Box caps.** CenterPoint keeps 100 boxes per head and our ensembling code 200 per frame by default, which cut the low-score tail of every class (the pedestrian head was full on every V2X-Real frame). We raised the caps to 600 to 1,000 boxes per member and 1,000 for every later stage, worth about +0.003 combined on each track.

## VII. Experiments

### A. Validation protocol

Labeled real frames served for validation and model selection; Section VIII summarizes the use of real data. On LUMPI we built four sets from the public measurements 4 and 6 (the test frames come from measurement 4), each frame at least 300 frames from every forbidden id: M4 (152 frames), M6 (102 frames) and two rare-class sets M4r (76) and M6r (110). Frames are cropped and subsampled to 50,000 points exactly like the test frames, and a candidate is adopted only if it improves all four sets. On V2X-Real we used the 359 infrastructure-LiDAR frames of the public validation split, verified point by point to share no frame with the test set, in two halves of three scenes each; a change must improve both halves, and site maps are built leave-one-scene-out.

The final rounds added in-window frames: non-forbidden frames recorded in the same sessions as the test frames. On LUMPI, M4w holds 100 frames of measurement 4 inside the test window, each at least 3 frames from every forbidden id. On V2X-Real, we took 264 primary-LiDAR frames from 22 recording groups of the public training archives and split them by group into two halves. Fifteen of these groups share a sequence with forbidden frames; the forbidden frames and their counterparts from a second LiDAR were excluded by path. Because these frames are adjacent in time to the test frames, we allowed only class-level rules and wrote every candidate and its adoption threshold down before evaluating it. No original validation set (the four LUMPI sets and the 359 V2X-Real frames above) was allowed to lose more than 0.005. Five adoptions deviated from this protocol; they are marked † in Sections VII-D and VII-E.

### B. Final results

Table III gives the official results and final ranks of our submissions: first on LUMPI and fifth on V2X-Real. The realism score is the mean normalized realism implied by the official combined score and mAP (official EMD 1.04 and 0.95, FPD 0.29 and 0.27 on LUMPI and V2X-Real). Density matching of the realism frames is the largest single gain (+0.14 combined on LUMPI); all other gains are in detection, where the site priors, the point-feature rules and the point-fitted heights each moved the score by 0.005 to 0.012 on both tracks. Sections VII-D and VII-E start from submissions scoring 0.5278 (LUMPI) and 0.7859 (V2X-Real), which already included member selection, rotation TTA (Section V) and small fusion and score changes.

TABLE III

Official results of the final ranked submissions

| Track | Submission | Combined | 3D mAP@0.5 | Realism | Rank |
|---|---|---|---|---|---|
| LUMPI | 1052645 | 0.5541 | 0.2047 | 0.947 | 1 |
| V2X-Real | 1054488 | 0.8102 | 0.5013 | 0.951 | 5 |

### C. Ablation: what transferred from validation to the test set

Table IV is an ablation of the components of our final submissions: for each, the gain on validation and on the test set when it was added. Gains are AP of the class named unless marked otherwise; test gains are from the development leaderboard except in the last two rows (final leaderboard).

On both tracks, the components that remove a systematic sim-to-real difference (coordinates, density, box convention,

TABLE IV
ABLATION: GAIN OF EACH COMPONENT ON VALIDATION AND ON THE TEST SET

| Component | Track, class | Validation | Test |
|---|---|---|---|
| Per-cell density matching | LUMPI, realism | 0.59 → 0.95 | combined +0.14 |
| Car z offset and real median size | LUMPI Car | +0.041 | +0.027 |
| Site priors | LUMPI Person, Bicycle | +0.086, +0.092 | +0.041, +0.022 |
| Two-wheeler extent rule | LUMPI Bicycle | +0.023 | +0.018 |
| Van, Bus and Truck copies into Car | LUMPI Car | +0.106 | +0.034 |
| Two clutter Person members | LUMPI Person | +0.062 (first) | +0.009, +0.001 |
| Truck size calibration | V2X-Real truck | +0.17 | +0.076 |
| Point-feature rules | V2X-Real pedestrian | +0.038 | +0.028 |
| Crowd model | V2X-Real pedestrian | +0.019 | +0.025 |
| Four dropout vehicle members | V2X-Real vehicle | +0.008 to +0.025 | +0.001 to +0.010 |
| Four z-shift truck members (dropout lineage) | V2X-Real truck | +0.006 to +0.040 | +0.007 to +0.057 |
| Near-range z ramp | V2X-Real pedestrian | +0.003 | +0.010 |
| Final round on in-window frames (VII-D) | LUMPI, mAP | +0.034 | +0.029 |
| Final rounds on in-window frames (VII-E) | V2X-Real, mAP | +0.051 | +0.028 |

static structures, point statistics of real objects) all gained on test, typically a quarter to three quarters of their validation gain (median about two thirds). Choices that fitted the few physical objects of a validation set passed validation but lost on test and are not part of the final submissions: a LUMPI Truck length refit gained 0.096 Truck AP on validation and lost 0.021 on test, and extra Person members lost up to 0.025 Person AP. Classes with 25 and 27 test boxes (Bus, Motorcycle; counts reported by the development leaderboard) are too small for either validation or the leaderboard to rank reliably, so in the second half of the development phase every adopted package was also confirmed as a whole on the leaderboard.

### D. Final LUMPI round on in-window frames

We first evaluated the 0.5278 package on M4w. Scored against all labeled boxes it reaches mAP 0.172, against 0.176 on the leaderboard, and no class differs by more than 0.08 from its test AP on the development leaderboard; M4w therefore ranks like the test set, while the original validation sets did not (Car AP 0.88 there, 0.65 on test). Restricted to labels that contain at least one point, the same predictions score 0.198 on M4w: 21% of its Car, 36% of its Truck and 20% of its Van labels contain no point of the cropped clouds, so no detector can find them.

Box-size statistics of the M4w labels explained three losses. One third of the Person labels are bike-sized boxes (1.97 × 0.81 × 1.86 m: a cyclist labeled as Person); 13% of the Truck labels are 8.1 × 5.6 × 3.2 m boxes, wider than any box the detector can output; and half of the Van labels are car-sized (3.7 × 1.8 × 1.6 m). The first two led to the rules of Section VI; the third has no geometric remedy. Table V shows the steps on M4w.

TABLE V
LUMPI VALIDATION ON M4W: AP PER CLASS AND MAP

| M4w, all labeled boxes | Pers. | Car | Bic. | Mot. | Bus | Truck | Van | mAP |
|---|---|---|---|---|---|---|---|---|
| Package on the leaderboard (0.5278) | 0.308 | 0.653 | 0.114 | 0.033 | 0.002 | 0.037 | 0.057 | 0.172 |
| + rules (i)–(iii) of Section VI | 0.359 | 0.653 | 0.144 | 0.048 | 0.002 | 0.058 | 0.057 | 0.189 |
| + selected Person members, extra Car runs | 0.371 | 0.658 | 0.144 | 0.048 | 0.002 | 0.058 | 0.057 | 0.191 |
| + Truck → Van copy | 0.371 | 0.658 | 0.144 | 0.048 | 0.002 | 0.058 | 0.091 | 0.196 |
| + Car from real-size members | 0.371 | 0.674 | 0.144 | 0.048 | 0.002 | 0.057 | 0.090 | 0.198 |
| + Bicycle from real-size members (final) | 0.371 | 0.674 | 0.202 | 0.048 | 0.002 | 0.057 | 0.090 | 0.206 |

Candidates were decided one at a time, in an order and against thresholds fixed in advance (typically +0.01 AP for the affected classes). Adopted: the two-wheeler refit; the rider-as-Person output, with its score mapping chosen among four variants; the wide-object box; six of eleven candidate Person members, chosen by single-member AP on M4w; the Car boxes of the Person members as additional Car runs; and the Truck → Van copy, which also passed the four original LUMPI sets (−0.001 to +0.007).

In a second step the Car members were replaced by members trained on real-size cars (Section IV-D); of three compositions fixed in advance, the one using only real-size members was the best (Car +0.016), and on the four original LUMPI sets its Car output gained 0.05 to 0.09. Taking Bicycle and Motorcycle from the same members raised Bicycle by 0.058 but lowered Person and Motorcycle by 0.012 and 0.010, more than the 0.005 allowed, so this candidate failed as written. We then took only the Bicycle class from it.† Because this variant was defined after an interim result, it also had to pass a stricter guard, which it did: the Bicycle AP of these members' ensemble is higher on all four original LUMPI sets (+0.002 to +0.076).

On M4w the gain over the 0.5278 package is +0.034 mAP. On interleaved halves of M4w it is +0.033 and +0.036 mAP; on the first and second half of the window +0.046 and +0.017, because the gains of the Person members, the Van copy and the Bicycle source depend on what is in the scene. On the leaderboard the package scored 0.5541 combined, against 0.5278: test mAP rose from 0.176 to 0.205 (+0.029, about 85% of the gain on M4w and inside the range of the splits), and the in-window calibration of the realism frames added about 0.002.

### *E. Final V2X-Real rounds on in-window frames*

In the first round we re-selected each class's members greedily on the first half of the 264 in-window frames. A candidate had to gain at least 0.010 class AP on the second half, not lose over all 264 frames, and lose at most 0.005 on the 359 original validation frames; where several candidates passed, the one with the best first-half AP was taken. The truck re-selection passed (second-half truck AP 0.503 to 0.546). Three further candidates were adopted although each narrowly missed one threshold: † the vehicle re-selection (second-half gain 0.0093), the pedestrian re-selection (loss of 0.0054 on the original validation frames) and rider-sized boxes from the rider-trained member of Section IV-B (second-half gain 0.0096), appended at their rider size. The two pedestrian changes together lowered pedestrian AP on the original validation frames by 0.010. Validation mAP on the 264 frames rose from 0.572 to 0.598.

The second round adopted a truck fusion threshold of 0.6, the two real-size truck members of Section IV-D with the heading consensus of Section VI, and a pedestrian set with re-selected members and score parameters. Second-half truck AP rose from 0.546 to 0.665, and truck AP on the original validation frames from 0.872 to 0.910. The pedestrian change † passed its threshold by 0.0005 but was composed after a related search result had been seen. Validation mAP on the 264 frames rose from 0.598 to 0.623. Over both rounds the combined test score rose from 0.7859 to 0.8102 (3D mAP 0.473 to 0.501).

Two diagnostics on the in-window frames locate the V2X-Real gap. First, most of it is vertical and concerns pedestrians: on the 264 frames the ratio of 3D to BEV AP is 0.91 for vehicles and 0.90 for trucks but 0.73 for pedestrians, and our pedestrian boxes follow the local ground closely while the labels scatter around it. Second, riders, which count as pedestrians, make up 7% of the pedestrian labels and were almost never matched by walker-sized boxes; this motivated the rider boxes of the first round.

## VIII. Use of real data and code

**Use of real data.** Models were trained only on UrbanTwin frames and the procedural objects of Section IV; no real points, labels or crops entered training or ground-truth sampling. Labeled non-forbidden real frames were used for validation, for the selection of models, ensemble members and post-processing rules, and for statistics: box sizes, offsets between predictions and labels, and point counts of real objects by range. Unlabeled non-forbidden point clouds gave the sensor positions, the ground height, the twin-to-real offset, the density model of the realism frames and the site maps of Section VI. No forbidden frame was used for training, validation, tuning or label inspection: LUMPI frames keep at least 300 frames from every forbidden id (3 frames for M4w), and on V2X-Real the forbidden frames and their counterparts from the second LiDAR were excluded by path before decoding. The test frames were read only by the inference pipeline, by the frame-overlap check of Section VII-A and by label-free diagnostics of their points and of the detector outputs (ground height, box counts and sizes, score ranges).

**Reproducibility.** The pipeline is an OpenPCDet fork with a UrbanTwin dataset module (synthetic pedestrians, riders, clutter, point dropout, z shift, box occlusion, real-size rebuilding), stand-alone tools for density matching, ensembling, test-time augmentation, calibration, priors, rules and packaging, and one script per training run and per package that records every setting. Code and instructions to reproduce both submissions: https://github.com/dipper13579/urbantwin-sim2real-lidar. All training ran on a single 8 GB laptop GPU; a base model takes about 2.5 hours and a fine-tune 40 minutes to 2.5 hours.

## IX. Limitations and conclusion

**Limitations.** Bus (25 test boxes) and Motorcycle (27) stay near zero AP on LUMPI, and neither class can be validated (about half of the M4w Bus labels belong to one car-sized object). Truck validation is driven by a handful of physical vehicles and did not transfer on LUMPI. The final rounds select rules on frames adjacent in time to the test frames: they measure this recording window, not generalization to other scenes, and the V2X-Real rounds reused the same frames for many candidates, so their validation gains overstate the test gain. On V2X-Real the dominant remaining loss is the vertical fit of pedestrians. We used one detector family (CenterPoint, xyz input; one member with a second stage) and one laptop GPU, so larger backbones and multi-architecture ensembles were not explored.

**Take-aways.** Six findings should carry over to other detectors trained on digital twins.

1. *Measure the twin before tuning the detector.* A per-class table of twin-versus-real size and points per range (Table II) located the largest gaps: twin cars are 15% shorter with half the points, and twin trucks 2.6 m shorter. Real labels are needed only as statistics.
2. *Correct a gap at the source.* A post-processing fix that works points to a gap in the data. Rescaling predicted trucks was the largest per-class test gain of the development phase (+0.076 truck AP), and retraining on real-size trucks, the source-level version of the same fix, gave the largest single gain of the final V2X-Real rounds.
3. *Synthesize what the twin lacks, negatives included.* On the V2X-Real validation frames, two thirds of the confident pedestrian boxes sat on street clutter that the twin does not contain.
4. *Pooled realism metrics measure occupancy.* Matching the per-cell density of unlabeled real frames reaches the real-versus-real ceiling without modeling a single object, so such scores say little about object fidelity.

5. *A fixed sensor gives label-free priors.* Detections and occupancy that recur across unlabeled frames mark static structures; down-weighting them gave +0.041 Person AP on test.
6. *Expect validation gains to shrink, and validate on frames that rank like the test set.* Corrections of systematic differences typically kept a quarter to three quarters of their validation gain. In-window LUMPI frames ranked the classes like the test set, and the final LUMPI round validated on them kept about 85% of its gain; the original validation sets did not (Car AP 0.88 there, 0.65 on test).

**Conclusion.** Correcting the digital twin rather than the detector produced most of our score: first place on LUMPI (0.5541) and fifth place on V2X-Real (0.8102), with CenterPoint models trained on synthetic data on one laptop GPU.

## Acknowledgment

We thank the organizers of the UrbanTwin Sim2Real LiDAR Challenge for the data, the evaluation and their clarifications, and Karma Laboratories, Inc. for sponsoring the conference registration fee. AI assistants (Anthropic's Claude and OpenAI's Codex) were used to write and run experiment code and to analyze results, and to help draft and edit the text; the authors checked all experiments, results and text and take full responsibility for the content.